\documentclass[letterpaper, 10 pt, conference]{ieeeconf}
\IEEEoverridecommandlockouts
\let\labelindent\undefined
\usepackage{times}
\usepackage{diagbox}
\usepackage{pstool}

\usepackage{multicol}
\usepackage{multirow}
\usepackage{verbatim}
\usepackage{amsmath}
\usepackage{amssymb}
\usepackage{amsfonts}
\usepackage{graphicx}
\usepackage{pstool}
\usepackage{breqn}
\usepackage{algorithm}
\usepackage[noend]{algpseudocode}
\usepackage[bookmarks=false]{hyperref}
\usepackage{paralist}
\usepackage{overpic}
\usepackage{xcolor,varwidth}
\usepackage{subcaption}
\usepackage{siunitx}

\let\eqref\undefined

\newcommand{\figref}[1]{Fig.~\ref{fig:#1}}
\newcommand{\algoref}[1]{Algorithm~\ref{algo:#1}}
\newcommand{\secref}[1]{Section~\ref{sec:#1}}
\newcommand{\tabref}[1]{Table~\ref{tab:#1}}
\newcommand{\eqref}[1]{Eq.~\ref{eq:#1}}

\newcommand{\figlabel}[1]{\label{fig:#1}}
\newcommand{\algolabel}[1]{\label{algo:#1}}
\newcommand{\seclabel}[1]{\label{sec:#1}}
\newcommand{\tablabel}[1]{\label{tab:#1}}
\newcommand{\eqlabel}[1]{\label{eq:#1}}

\newcommand{\U}[0]{\mathbf{u}}
\newcommand{\X}[0]{\mathbf{x}}
\newcommand{\V}[0]{\mathbf{v}}
\newcommand{\Q}[0]{\mathbf{q}}
\newcommand{\p}[0]{\mathbf{p}}
\newcommand{\initial}[1]{{#1}^\mathcal{I}}
\newcommand{\final}[1]{{#1}^\mathcal{F}}
\newcommand{\alphas}{\vec{\boldsymbol{\alpha}}}

\newcommand{\axisfont}{\fontsize{6}{6}}

\DeclareMathOperator*{\argmax}{arg\,max}
\DeclareMathOperator*{\argmin}{arg\,min}
\DeclareMathOperator*{\sign}{sign}

\DeclareMathOperator*{\atan2}{atan2}

\makeatletter
\let\OldStatex\Statex
\renewcommand{\Statex}[1][3]{%
  \setlength\@tempdima{\algorithmicindent}%
  \OldStatex\hskip\dimexpr#1\@tempdima\relax}
\makeatother

\title{\LARGE \bf
A Real-Time Solver For Time-Optimal Control Of \\
Omnidirectional Robots with Bounded Acceleration
}

\author{David Balaban, Alexander Fischer, and Joydeep Biswas$^{1}$% <-this % stops a space
  \thanks{$^{1}$The authors are with the University of Massachusetts Amherst,
USA. Email: {\tt\small \{dblaban, afischer, jbiswas\}@umass.edu}. This work is
supported in part by AFRL and DARPA under agreement \#FA8750-16-2-0042, and NSF
grant IIS-1724101.}%
}

\begin{document}

\maketitle
\thispagestyle{empty}
\pagestyle{empty}

\begin{abstract}
We are interested in the problem of time-optimal control of omnidirectional
robots with bounded acceleration (TOC-ORBA). While there exist approximate
solutions for such problems, and exact solutions with unbounded acceleration,
exact solvers to the TOC-ORBA problem have remained elusive until now. In this
paper, we present a real-time solver for true time-optimal control of
omnidirectional robots with bounded acceleration. We first derive the general
parameterized form of the solution to the TOC-ORBA problem by application of
Pontryagin's maximum principle. We then frame the boundary value problem of
TOC-ORBA as an optimization problem over the parameterized control space. To
overcome local minima and poor initial guesses to the optimization problem, we
introduce a two-stage optimal control solver (TSOCS): The first stage
computes an upper bound to the total time for the TOC-ORBA problem and holds
the time constant while optimizing the parameters of the trajectory to approach
the boundary value conditions. The second stage uses the parameters found
by the first stage, and relaxes the constraint on the total time to solve for
the parameters of the complete TOC-ORBA problem. Furthermore, we implement TSOCS as a
closed loop controller to overcome actuation errors on real robots in
real-time. We empirically demonstrate the effectiveness of TSOCS in
simulation and on real robots, showing that 1) it runs in real time, generating
solutions in less than 0.5ms on average; 2) it generates faster trajectories
compared to an approximate solver; and 3) it is able to solve TOC-ORBA problems
with non-zero final velocities that were previously unsolvable in real-time.

\end{abstract}

\IEEEpeerreviewmaketitle

\section{Introduction}
Omnidirectional robots find use in a variety of domains where high
maneuverability is important, such as indoor service mobile
robots~\cite{biswas20161000km}, robot soccer~\cite{mendoza2016selective}, and
warehouse automation~\cite{rohrig2010localization}. Such robots rely on one of
several wheel
designs~\cite{pin1994new} to decouple the kinematic constraints along the three
degrees of freedom, two for translation and one for rotation, necessary for
motion on a plane. The dynamics of the omnidirectional drive is constrained by
the specific layout of the wheel configuration, and the drive motor
characteristics. Owing to the complexity of modelling the angle-dependent
dynamic constraints imposed by every motor, omnidirectional drive systems may be
simplified as having acceleration and velocity that are bounded uniformly in all directions~\cite{kalmar2004near}.
However, despite such simplifications, the solution to the time-optimal control
of omnidirectional robots with bounded acceleration remains elusive: while there
have been time-optimal solutions for omnidirectional robots with unbounded
acceleration~\cite{wang2012analytical}, and approximate solutions to
time-optimal control with bounded acceleration and bounded
velocity~\cite{kalmar2004near}, until now, there has been no solution to the
true time-optimal control of omnidirectional robots with bounded acceleration.

In this paper, we present an algorithm for true time-optimal control of an
omnidirectional robot with bounded acceleration (TOC-ORBA) but unbounded velocity. By applying
Pontryagin's maximum principle to the dynamics of the omnidirectional robot with
bounded acceleration, we derive the necessary conditions for the solution to the
TOC-ORBA problem, and the adjoint variables of the problem. By analyzing the
adjoint-space formulation, we derive the parametric form of a solution to
the TOC-ORBA problem. Given a specific problem with initial and final
conditions, we frame the TOC-ORBA problem as a boundary-value problem,  to solve
for the parameters of the optimal trajectory that satisfy the initial and final
conditions.

However, this boundary value problem cannot be solved analytically, and direct
application of optimization solvers result in poor convergence rates and
sub-optimal local minima~\cite{Pifko2008}. We therefore decouple the full
solution into two stages to overcome local minima. In the first stage, we first
compute the upper bound for the total time of the TOC-ORBA problem by
decomposing the problem into three linear motion segments that are always
solvable analytically, but which will be sub-optimal. We hold the total
trajectory time constant at this upper bound, and solve for the parameters of
the trajectory to approach the boundary value conditions. Since the total time
is held constant in the first stage, it may not be able to exactly satisfy the
boundary conditions, but will typically find the correct shape of the optimal
trajectory. In the second stage, we use the first stage solution as an initial
guess, relax the constraint on the total time, and solve for the parameters
that exactly satisfy the boundary conditions. Thus, our two-stage optimal
control solver (TSOCS) is able to compute exact solutions to the TOC-ORBA
problem.

We further show how TSOCS can be modified for practical use in an
iterative closed-loop controller for real robots with inevitable actuation
errors. We demonstrate the closed-loop TSOCS controller running in real time
at 60Hz on real robots, taking 0.5ms on average to solve the TOC-ORBA problem.
We present results from experiments over extensive samples from the TOC-ORBA
problem space with simulated noisy actuation, as well as on real robots.
The evaluation of TSOCS shows that it outperforms existing approximate
solvers~\cite{kalmar2004near}, and that it can solve for, and execute
time-optimal trajectories with non-zero final velocities, which were previously
not solvable. We further demonstrate how varying a hyperparameter in the TSOCS
controller can be used to trade-off accuracy in final location vs. final
velocity.

\section{Background and Related Work}
\seclabel{background}

Omnidirectional robots, also referred to as holonomic drive robots, are
governed by the following system of ordinary differential equations:
\begin{align}
\frac{dx_1}{dt} = x_3, \quad
\frac{dx_2}{dt} = x_4, \quad
\frac{dx_3}{dt} = u_1, \quad
\frac{dx_4}{dt} = u_2
\eqlabel{dynamics}
\end{align}
where $x_1,\ x_2$ are the Cartesian coordinates of the robot's position, $x_3,
\ x_4$ are the robot's velocity and $ u_1, \ u_2 $ are the accelerations along the $x_1$ and $x_2$ directions respectively. The
velocity and acceleration of the robot are limited by the maximum speed and torque of the driving motors. While such limits are dependent on the number of omnidirectional wheels on the robot and their orientations~\cite{kalmar2004near}, we consider orientation-independent bounds as the minimum possible bound in any orientation~\cite{kalmar2004near}:
\begin{align}
(u_1^2 + u_2^2)^{\frac{1}{2}}  \leq u_{\mathrm{max}}, \quad
(x_3^2 + x_4^2)^{\frac{1}{2}} \leq v_{\mathrm{\mathrm{max}}}
\eqlabel{bounds}
\end{align}
The initial state of the system is denoted by $\initial x_i$, for all state
coordinates $i\in[1,n]$, and the final state by $\final{x_i},i\in[1,n]$. The
objective of the time-optimal control (TOC) problem is to find an optimal control input function
$\U^*(t)$ among all admissible control functions which drives the
system along an optimal trajectory $\X^*(t)$, $0\leq t\leq T^*$ such that $\X(T^*)=\final\X$ and $T^*$ is minimized. Here we outline common
strategies for solving optimal robotic control problems, describe previous work
done on omnidirectional robots, and explain the contributions of this paper.

There are three main strategies employed to solve optimal control problems of this type. The first uses system dynamics to find an analytical solution with Pontryagin's Maximum Principle (PMP)~\cite{pontryagin1987mathematical}. This approach has been used in the control of two-wheeled robots~\cite{renaud1997minimum}, two-legged robots~\cite{pontryagin1987mathematical} and robots with a trailing body~\cite{chyba1999time}. A second approach solves the system numerically as an optimization problem--controlling the joint angles of a robot along a predefined path can be solved as a convex optimization~\cite{verscheure2009time}. A recurrent neural network can be used to satisfy Karush-Kuhn-Tucker conditions~\cite{kuhn1951} on a dynamic structure known as a stewart structure~\cite{mohammed2016dynamic}. The third approach is to discretize space and/or time and search for an optimal solution. This approach can be used to find time optimal paths that avoid collisions between coordinating robots with predefined paths~\cite{altche2016time}. The discretization method is common for path planning~\cite{davoodi2013multi}.

No time-optimal solution has been found for omnidirectional robots which
accounts for constraints on velocity and acceleration. However, there have been successful solutions which either relax the constraints on the problem or find approximate solutions. A linearized kinematic model~\cite{li2009motion} and non-linear controller~\cite{penaloza2015motion} can enforce wheel velocity constraints. An analytical near-time-optimal control (NTOC)~\cite{kalmar2004near} accounts for
both the acceleration and velocity constraints; however, if the wheel velocity constraints are considered, but
acceleration is left unbounded, time optimal solutions follow certain classes
of optimal trajectories~\cite{balkcom2006time} with analytical solutions~\cite{wang2012analytical}. Previous work found the solution form to time-optimal control of omnidirectional robots with bounded acceleration (TOC-ORBA) and unbounded velocity~\cite{Pifko2008}, and we build upon this work
and introduce a real-time solver capable of reliably solving the TOC-ORBA problem. With this solver, an omnidirectional robot can reach goal states with non-zero velocity, which the previous state-of-the-art NTOC~\cite{kalmar2004near} could not solve.

\section{Solution Form For TOC-ORBA}
Pontryagin's maximum principle (PMP)~\cite{pontryagin1987mathematical} provides
necessary
conditions for the optimal control $\U^*(t)$ to minimize $T$. PMP for TOC
problems is stated in terms of the Hamiltonian of the
system,
\begin{align}
    H(\Psi,\X,\U) = -1 + \sum_{i=1}^n \psi_i \frac{dx_i}{dt},
\end{align}
where $\Psi=\psi_1,\ldots,\psi_n$ are the adjoint variables of the system constrained
by the following ODEs:
\begin{align}
\frac{d\psi_i}{dt} = \frac{\partial H}{\partial x_i} \qquad
\frac{dx_i}{dt} = \frac{\partial H}{\partial \psi_i}
\eqlabel{adjoint_vars}
\end{align}
PMP states that the optimal control $\U^*$ must maximize the Hamiltonian among
all admissible control inputs $\U$ at every time step $t$: $H(\U^*) =
\argmax_\U H(\U)$. For omnidirectional robots with bounded acceleration, the Hamiltonian is
\begin{align}
H(\Psi,\X,\U) = -1 + \psi_1 x_3 + \psi_2 x_4 + \psi_3 u_1 + \psi_4 u_2,
\end{align}
and the adjoint variables $\Psi=\psi_1,\psi_2,\psi_3,\psi_4$, constrained by
\eqref{adjoint_vars}, must satisfy:
$
\frac{d\psi_1}{dt} = 0;
\frac{d\psi_2}{dt} = 0;
\frac{d\psi_3}{dt} = \psi_1;
\frac{d\psi_4}{dt} = \psi_2
$
Thus, the adjoint variables $\psi_1,\psi_2$ are constant in time, and
$\psi_3,\psi_4$ vary linearly with time, yielding
\begin{align}
\psi_1 &= \alpha_1 \qquad \qquad
\psi_2 = \alpha_2 \notag \\
\psi_3 &= \alpha_1 t + \alpha_3 \qquad
\psi_4 = \alpha_2 t + \alpha_4.
\eqlabel{adjoint_form}
\end{align}
The constants $\alphas=\{\alpha_1,\alpha_2,\alpha_3,\alpha_4\}$ in the
expressions for
$\Psi$ depend on the boundary conditions $\X(0),\X(T)$ of the problem. We
reformulate the control variable as
\begin{align}
u_1 = a \cos \theta, \qquad
u_2 = a \sin \theta.
\eqlabel{reform_control}
\end{align}
The Hamiltonian of the reformulated problem is
thus
\begin{align}
H = -1 + \psi_1 x_3 + \psi_2 x_4  +
a(\psi_3\cos \theta + \psi_4\sin\theta).
\end{align}
Since the Hamiltonian is linear with respect to $a$, from PMP, the time-optimal
radial acceleration $a^*$ is given by,
\begin{align}
a^* = u_{\mathrm{max}}\sign(\psi_3\cos \theta + \psi_4\sin\theta),
\eqlabel{property1}
\end{align}
This expression gives us the first insight to the problem: \\
\textbf{Property 1}: \emph{The magnitude of the time-optimal acceleration is
  always equal to the acceleration bound of the robot.}

To find the optimal values of $\theta$, we find the extrema of the Hamiltonian
with respect to $\theta$:
\begin{align}
\left.\frac{\partial H}{\partial \theta}\right|_{\theta = \theta^*} = 0 \
\implies \theta^* = \atan2(\psi_4, \psi_3)
\eqlabel{property2}
\end{align}
This expression gives us the second insight to the problem:\\
\textbf{Property 2}: \emph{The direction of the time-optimal acceleration is parallel to the line joining the origin and $(\psi_3,\psi_4)$.}

$\theta^*$ will always point towards $(\psi_3,\psi_4)$. However, it could be possible that $a^*$ is negative, in which case the direction of the time-optimal acceleration would point away from $(\psi_3,\psi_4)$. Given $\theta^*$ we can find expressions for $\sin\theta^*$ and $\cos\theta^*$,
then use \eqref{property1} to get the third property:
\begin{align}
a^* &= u_{\mathrm{max}}\sign\left(\frac{\psi_3^2 + \psi_4^2}{\sqrt{\psi_3^2+\psi_4^2}}\right)=u_{\mathrm{max}}
\eqlabel{property3}
\end{align}
\textbf{Property 3}: \emph{$a^*=u_{\mathrm{max}}$ is always positive, so the acceleration always points in the direction of the point $(\psi_3,\psi_4)$.}

The adjoint variables make a
parametric line, with linear dependence on time (\eqref{adjoint_form}). The acceleration always lies
on a circle with radius $u_{\mathrm{max}}$, with its direction pointing towards the corresponding point on the adjoint line.

For the remainder of this section, we let $u_{\mathrm{max}}=1$ without loss of generality for ease of readability. From \eqref{reform_control} and \eqref{property3} we get the form of the
optimal control in Cartesian Coordinates as a function of the adjoint variables:
\begin{align}
u_1^* = \frac{\psi_3}{\sqrt{\psi_3^2 + \psi_4^2}}, \
u_2^* = \frac{\psi_4}{\sqrt{\psi_3^2 + \psi_4^2}},
\eqlabel{optimal_form}
\end{align}
We choose coordinate axes such that the robot always starts from the origin
with an initial velocity $\initial v_1$ and $\initial v_2$. For ease of readability, we
rename the variables $x_3$ and $x_4$ as $v_1$ and $v_2$ to reflect their role
as the velocity in the $x_1$ and $x_2$ directions. We also use boldface vector
notation, $\X=[x_1~x_2]^T, \V=[v_1~v_2]^T, \U=[u_1~u_2]^T$.
We find the velocity $\V$ by integrating the control function over time and the position $\X$ by integrating the resulting velocity over time.

\subsection{Solution to TOC-ORBA as a Boundary-Value Problem}

From \eqref{optimal_form}, the optimal control is symmetric in the
adjoint variables, therefore the time-optimal solution to $u_1$ and $u_2$ will also be correspondingly symmetric. We thus derive the expressions for the position $x_1$ and velocity $v_1$, noting that the expressions for $x_2$ and $v_2$ will be symmetric in form with a change of variable from $\alpha_1,\alpha_3$ to $\alpha_2,\alpha_4$. Let $\p = \begin{bmatrix} \alpha_3 \\ \alpha_4 \end{bmatrix}$, $\Q = \begin{bmatrix} \alpha_1 \\ \alpha_2 \end{bmatrix}$, and make the following abbreviations:
\begin{align}
h_1 &= \sqrt{\Psi_3^2 + \Psi_4^2} \qquad \quad
h_2 = h_1 ||\Q|| + ||\Q||t^2 + \p \cdot \Q \notag \\
h_3 &= ||\p||\ ||\Q||\ +\ \p \cdot \Q \quad
\gamma = \frac{h_2}{h_3}
\end{align}

The velocity and position can then be expressed as below:

\begin{align}
v_1&(t,\p,\Q) = \initial v_{1} + \alpha_1 \frac{h_1-||\p||}{||\Q||^2}+\alpha_2
\frac{\det(\p\ \Q)}{||\Q||^3}\ln(\gamma) \notag \\
x_1&(t,\p,\Q) = \initial x_{1} + \initial v_{1}t + \notag
\frac{\alpha_1}{2||\Q||^5} \Big(h_1\left(||\Q||\p \cdot \Q + t||\Q||^3\right) \notag \\
&+ ||\p \times \Q||^2\ln(\gamma)
 - ||\p||\left(||\Q||\p \cdot \Q + 2 t||\Q||^3\right)\Big) \notag \\
 &+ \frac{\alpha_2 \det(\p\ \Q)}{||\Q||^3} \left(\ln(\gamma)\left(t +
 \frac{\p \cdot \Q}{||\Q||^2}\right)
 - \frac{h_1 - ||\p||}{||\Q||}\right)
\end{align}
We now have an expression for each of the four
state variables $(x_1(t), x_2(t), v_1(t), v_2(t))$
describing the optimal trajectory given values for $\alphas$ at any
desired time $t: [0, T]$. To solve the TOC-ORBA problem we therefore must find
the parameters $\alphas, T$ such that they satisfy the boundary
conditions: $\X(\alphas,T) = \final{\X}, \V(\alphas,T)=\final{\V}$ for given values of
$\final{\X}$ and $\final{\V}$. By design, the initial conditions
$\X(\alphas,0)=\initial{\X}=0, \V(\alphas,0)=\initial{\V}$ are already satisfied for
any given values of $\initial{\V}$. We can solve this system of
equations as a boundary value problem (BVP) with four non linear equations from
the final conditions constraining the five parameters.

\section{Solving TOC-ORBA by Nonlinear \\
Least Squares Optimization}
\seclabel{LSO}

TOC-ORBA can be solved by evaluating the nonlinear least squares
optimization problem
\begin{align}
\boldsymbol{\alphas^*}, T^* &= \argmin_{\alphas,T} F_{\mathrm{BV}}(\alphas, T), \\ \notag
F_{\mathrm{BV}}(\alphas, T) &= ||\final{\X} - \X(\alphas,T)||^2 + ||\final{\V} - \V(\alphas,T)||^2,
\eqlabel{tsocs-cost}
\end{align}
where $F_{\mathrm{BV}}$ is the cost function that penalizes control parameters that violate the boundary value constraints.

We derive an upper bound on the optimal time $T^*$ for the TOC-ORBA problem by decomposing the 2D motion control problem into the following sequence of 1D
problems: 1) First, the robot accelerates to rest if it has initial velocity. 2) Second, the robot moves to the point from which it can directly accelerate to the goal state (location and velocity) from rest. 3) Third, the robot accelerates directly to the goal state. The upper bound on $T^*$ is thus the sum of the time taken for all three steps, given by:
\begin{align}
T_{\mathrm{max}} = \frac{||\initial{\V}|| + ||\final{\V}||}{u_{\mathrm{max}}} +
2\sqrt{\left\lVert\frac{\final{\X}}{u_{\mathrm{max}}^2} - \frac{\final{\V}||\final{\V}|| +
\initial{\V}||\initial{\V}||}{2u_{\mathrm{max}}^3}\right\rVert}
\end{align}

For an initial guess of the parameters, we project the initial and goal velocities onto the displacement vector of the start and goal locations, and solve the corresponding one-dimensional time-optimal control problem. Regardless of the initialized parameters, non-linear least-squares optimization is not guaranteed to find a correct solution if the cost function contains local minima.

\begin{figure}
  \centering
  \begin{overpic}[width=\linewidth]{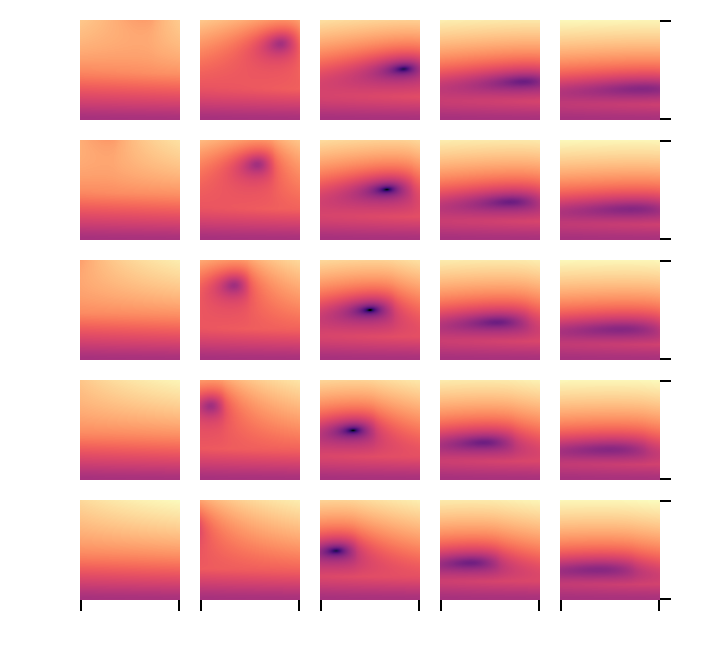}
  
    % left side
    \put(0,13.5){\axisfont$\alpha_2\!\! :0.99$}
    \put(0,30.5){\axisfont$\alpha_2\!\!:0.83$}
    \put(0,47.5){\axisfont$\alpha_2\!\!:0.66$}
    \put(0,64.5){\axisfont$\alpha_2\!\!:0.50$}
    \put(0,81.5){\axisfont$\alpha_2\!\!:0.33$}
    
    % top side
    \put(12.5,90){\axisfont$\alpha_1\!\!:1.90$}
    \put(29.5,90){\axisfont$\alpha_1\!\!:2.86$}
    \put(46.4,90){\axisfont$\alpha_1\!\!:3.81$}
    \put(63.4,90){\axisfont$\alpha_1\!\!:4.76$}
    \put(80.4,90){\axisfont$\alpha_1\!\!:5.71$}
    
    % bottom side
    \put(17,1){\axisfont$\alpha_4$}
    \put(34,1){\axisfont$\alpha_4$}
    \put(51,1){\axisfont$\alpha_4$}
    \put(68,1){\axisfont$\alpha_4$}
    \put(85,1){\axisfont$\alpha_4$}

    \put(10,3.5){\axisfont$-2.33$}
    \put(19,3.5){\axisfont$-0.78$}
    
    \put(27,3.45){\axisfont$-2.33$}
    \put(36,3.45){\axisfont$-0.78$}
    
    \put(44,3.4){\axisfont$-2.33$}
    \put(53,3.4){\axisfont$-0.78$}
    
    \put(61,3.35){\axisfont$-2.33$}
    \put(70,3.35){\axisfont$-0.78$}
    
    \put(78,3.3){\axisfont$-2.33$}
    \put(87,3.3){\axisfont$-0.78$}
    
    % right side
    \put(95,13.5){\axisfont$T$}
    \put(95,30.5){\axisfont$T$}
    \put(95,47.5){\axisfont$T$}
    \put(95,64.5){\axisfont$T$}
    \put(95,81.5){\axisfont$T$}

    \put(95,6.5){\axisfont$0.04$}
    \put(95,20.5){\axisfont$7.63$}
    
    \put(95,23.45){\axisfont$0.04$}
    \put(95,37.45){\axisfont$7.63$}
    
    \put(95,40.4){\axisfont$0.04$}
    \put(95,54.4){\axisfont$7.63$}
    
    \put(95,57.35){\axisfont$0.04$}
    \put(95,71.35){\axisfont$7.63$}
    
    \put(95,74.3){\axisfont$0.04$}
    \put(95,88.3){\axisfont$7.63$}
    
%    % left side
%    \put(-1,13.5){\fontsize{5}{6}$\alpha_2$}
%    \put(-1,30.5){\fontsize{5}{6}$\alpha_2$}
%    \put(-1,47.5){\fontsize{5}{6}$\alpha_2$}
%    \put(-1,64.5){\fontsize{5}{6}$\alpha_2$}
%    \put(-1,81.5){\fontsize{5}{6}$\alpha_2$}
%    
%    % top side
%    \put(11.5,90){\fontsize{5}{6}$\alpha_1$}
%    \put(28.5,90){\fontsize{5}{6}$\alpha_1$}
%    \put(45.5,90){\fontsize{5}{6}$\alpha_1$}
%    \put(62.5,90){\fontsize{5}{6}$\alpha_1$}
%    \put(79.5,90){\fontsize{5}{6}$\alpha_1$}
%    
%    % bottom side
%    \put(17,1){\fontsize{5}{6}$\alpha_4$}
%    \put(34,1){\fontsize{5}{6}$\alpha_4$}
%    \put(51,1){\fontsize{5}{6}$\alpha_4$}
%    \put(68,1){\fontsize{5}{6}$\alpha_4$}
%    \put(85,1){\fontsize{5}{6}$\alpha_4$}
%    
%    % right side
%    \put(95,13.5){\fontsize{5}{6}$T$}
%    \put(95,30.5){\fontsize{5}{6}$T$}
%    \put(95,47.5){\fontsize{5}{6}$T$}
%    \put(95,64.5){\fontsize{5}{6}$T$}
%    \put(95,81.5){\fontsize{5}{6}$T$}
  \end{overpic}
  \caption{Visualization of $F_{\mathrm{BV}}$ with $\alpha_3 = -6.81$. In each panel $\alpha_4$ varies along the $x$ axis and $T$ along the $y$ axis. Darker colors correspond to lower cost.}
  \vspace{-1em}
  \label{fig:fs2}
\end{figure}

\begin{figure}
  \vspace{0.5em}
  \centering
  \includegraphics[width=\linewidth]{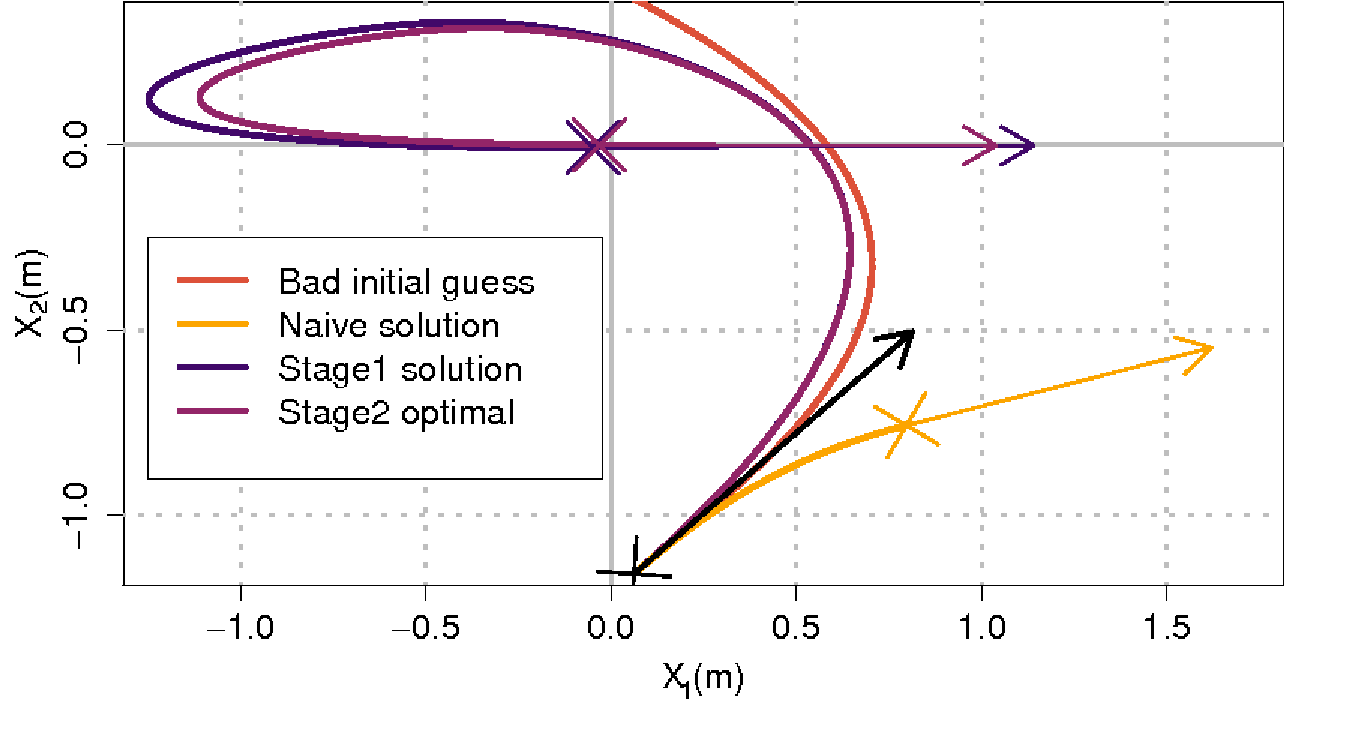}
  \vspace{-2.0em}
  \caption{An example TOC-ORBA problem: solutions found by the different
    solvers, compared to the optimal and initial paths given a poor initial
    guess; arrows show direction and relative magnitude of final velocity.}
  \label{fig:paths}
  \vspace{-2.0em}
\end{figure}

\figref{fs2} shows a visualization of $F_{\mathrm{BV}}$ for an example
TOC-ORBA problem. The global minimum is in the center of the center panel, and there is a local minimum in the area near $T=0$. \figref{paths} shows the path corresponding to the
local minimum, which was obtained by running the nonlinear least-squares solver from a random initial guess.


\subsection{Two-Stage Solver For TOC-ORBA}

To avoid the problem of local minima, we split the solver into two stages: stage
1 holds $T$ constant at the computed upper bound~(\secref{LSO}) and minimizes
the cost function $F_{\mathrm{BV}}$; and stage 2 takes the parameter set found
by stage 1 as an initial guess and minimizes $F_{\mathrm{BV}}$ further by
allowing $T$ to vary. We call this solution the two-stage optimal control solver
(TSOCS).  Despite the fact that stage 1 is unlikely to find a valid solution,
the parameter set found typically provides an initial guess to stage 2 within
its optimal basin of attraction. \figref{paths} shows the paths corresponding to
the local minimum and the global minimum found by TSOCS.

\begin{comment}
To demonstrate an example how resilient the 2 stage solver is to poor initialization,
we compare the two solvers solutions when an intentionally poor initial guess is given.
Fig. \ref{fig:paths} shows the paths each solver stage finds, as well as the
initial guess they started with. In the initial guess, the robot accelerates
towards $-\infty$ along the $x_1$-axis for 100 seconds and never changes
direction. The naive solver gets stuck in a local minimum as discussed in
\secref{minima}. With TSOCS, the first stage finds a
solution that ends at the correct position, but with a much higher velocity,
the second stage then corrects the path and finds the correct solution.
\end{comment}

\subsection{Iterative Closed-Loop Control}
\seclabel{iterative}
A real omnidirectional robot will have actuation errors, which will perturb it
from the desired trajectory.
To overcome such actuation errors, we present an iterative closed-loop
controller using TSOCS, which re-solves the TOC-ORBA problem
with new observations at every time-step.

\newcommand{\lineref}[1]{line~\ref{line:#1}}

\algoref{iterative-tsocs} lists IterativeTSOCS, which implements iterative
closed-loop control using TSOCS. Given the desired final location and velocity
$\final{\X}, \final{\V}$, IterativeTSOCS first computes the initial guess to
the optimal control parameters $\alphas, T$~(\lineref{stage1}). Next, it
executes the iterative control loop
until the robot reaches the desired final location and
velocity~(\lineref{loop-start}). At each time-step, IterativeTSOCS updates the
current state of the robot based on new observations~(\lineref{observe}), and
computes a new initial guess for the optimal control parameters by moving
the initial adjoint point forward along the adjoint line~(\lineref{update1}--%
\lineref{update2}) by the time-period $\Delta T$.

To prevent frequent backtracking due to small actuation errors, IterativeTSOCS
runs an variant of the TSOCS second stage solver
with a modified cost function $F_{\mathrm{it}}$,
\begin{align}
F_{\mathrm{it}} = &||\final{\X} - \X(T)||^2 + \beta||\final{\V} - \V(T)||^2 +
         k_1e^{k_2(T / T_e - \tau)}, \notag \\
\beta = &\max\left(1 - \frac{||\final{\V} - \initial{\V}||}{u_{\mathrm{max}} T_e},
\beta_{\mathrm{min}}\right).
\eqlabel{iterative-cost}
\end{align}
$F_{\mathrm{it}}$ modifies the boundary value cost function~(\eqref{tsocs-cost})
by including a regularization cost for the total time $T$, and a discount factor
$\beta$ for the velocity error. The time regularization term, $k_1e^{k_2(T / T_e - \tau)}$, grows large if $T$ becomes more than $\tau$ times $T_e$, the expected time based off the previous iteration's time. This prevents the robot from backtracking to correct for small actuation errors, which would take much more time that following the original trajectory. We chose $\tau=1.4$ in our experiments on real robots. The
discount factor $\beta$ for the velocity error allows IterativeTSOCS to reach the final
state with small errors in velocity while maintaining location accuracy, which would have been otherwise dynamically
infeasible for the robot to correct without backtracking. $\beta$ decreases as the problem becomes near one-dimensional, because TSOCS fails on near one-dimensional cases more frequently.

If any iteration fails to find path using the paremeter guess from the updated adjoint line, then IterativeTSOCS runs the full two-stage solver on that iteration, with the second stage using the cost function in \eqref{iterative-cost}. If that fails as well, then IterativeTSOCS resorts to following open loop control from the last successful parameter set found.

\vspace{-0.2cm}
\begin{algorithm}[htb]
\caption{Iterative TSOCS}
\algolabel{iterative-tsocs}
\small
\begin{algorithmic}[1]
\Procedure{IterativeTSOCS}{$\final{\X},\final{\V}$}
\State $\initial{\X}, \initial{\V} \gets$ Observe()
\State $\langle \alphas, T \rangle \gets$ Stage1($\initial{\X},\initial{\V},\final{\X},\final{\V}$)\label{line:stage1}
\While{$||\initial{\X}-\final{\X}||>\epsilon_\X \wedge ||\initial{\V}-\final{\V}||>\epsilon_\V \wedge T > \Delta T$} \label{line:loop-start}
\State $\langle \alphas_{\mathrm{n}}, T_{\mathrm{n}}, F_{\mathrm{it}} \rangle \gets$ Stage2It($\initial{\X},\initial{\V},\final{\X},\final{\V}, \alphas, T$)\label{line:stage2}
\If {$F_{\mathrm{it}} < \epsilon_{\text{cost}}$}
\State $\langle \alphas, T \rangle \gets \langle \alphas_{\mathrm{n}}, T_{\mathrm{n}} \rangle$
\State $T \gets T_{\mathrm{n}}$
\Else
\State $\langle \alphas_{\mathrm{n}}, T_{\mathrm{n}} \rangle \gets$ Stage1($\initial{\X},\initial{\V},\final{\X},\final{\V}$)
\State $\langle \alphas_{\mathrm{n}}, T_{\mathrm{n}}, F_{\mathrm{it}} \rangle \gets$ Stage2It($\initial{\X},\initial{\V},\final{\X},\final{\V}, \alphas, T$)
\If {$F_{\mathrm{it}} < \epsilon_{\text{cost}}$}
\State $\langle \alphas, T \rangle \gets \langle \alphas_{\mathrm{n}}, T_{\mathrm{n}} \rangle$
\EndIf
\EndIf
\State Execute($\alphas$)\label{line:execute}
\State $\alphas.\alpha_3 \gets \alphas.\alpha_3 + \alphas.\alpha_1\Delta T$\label{line:update1}
\State $\alphas.\alpha_4 \gets \alphas.\alpha_4 + \alphas.\alpha_2\Delta T$
\State $T \gets T - \Delta T$ \label{line:update2}
\State $\initial{\X}, \initial{\V} \gets$ Observe() \label{line:observe}
\EndWhile \label{line:loop-end}
\EndProcedure
\end{algorithmic}
\end{algorithm}
\vspace{-0.2cm}

After Stage2It recomputes the optimal control parameters $\alphas$
(\lineref{stage2}), the robot executes the
control for the time-step according to the updated parameters $\alphas$
(\lineref{execute}).

\section{Experimental Results} \seclabel{experimental}

\begin{figure}
\captionsetup[subfigure]{aboveskip=-1pt,belowskip=-4pt}
\centering
\begin{subfigure}[b]{0.49\linewidth}
\begin{overpic}[width=\linewidth,unit=1mm]{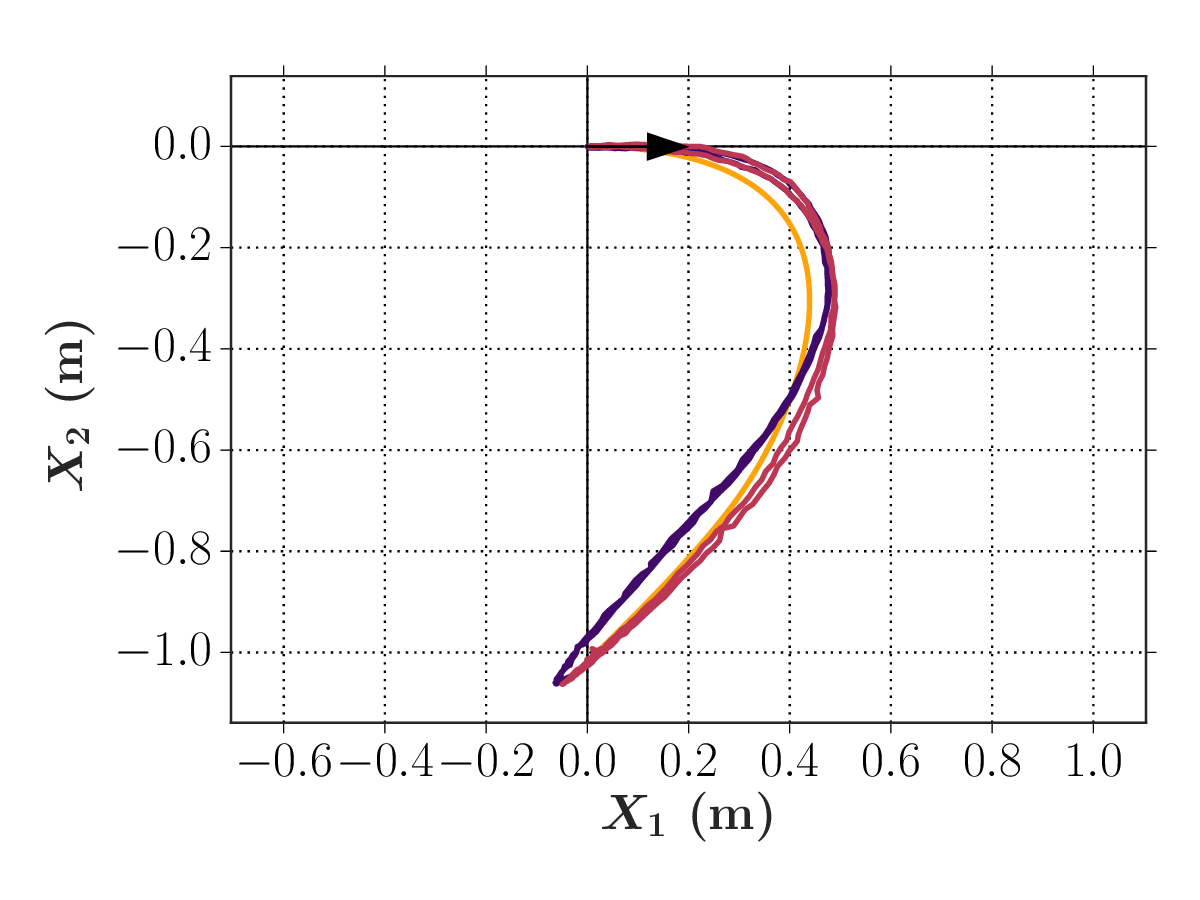}
\end{overpic}
\caption{Zero final velocity}
\end{subfigure}
\begin{subfigure}[b]{0.49\linewidth}
\begin{overpic}[width=\linewidth,unit=1mm]{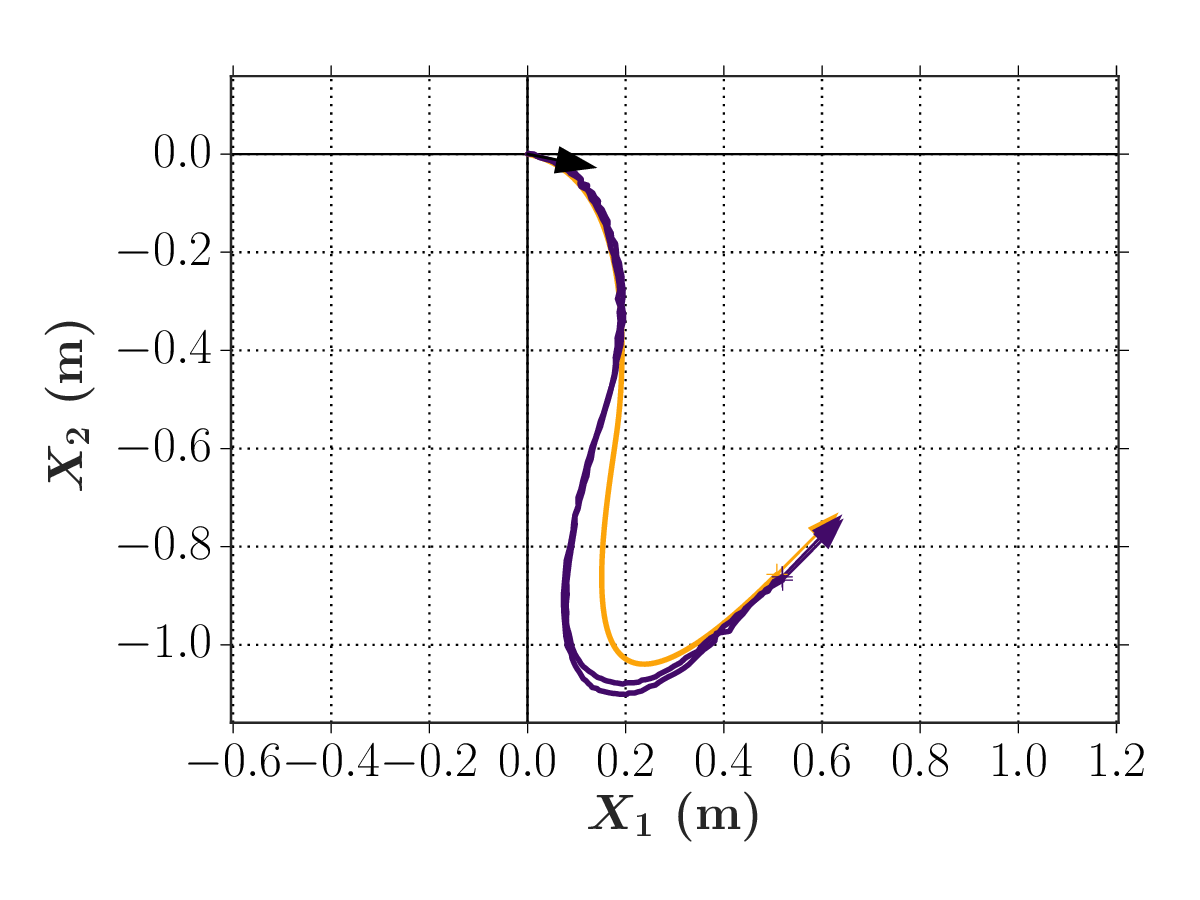}
\end{overpic}
\caption{Non-zero final velocity}
\end{subfigure}
\caption{Examples of TSOCS and NTOC run on real robots. Yellow is the optimal
path computed by TSOCS, blue is the path taken by real robots runnings TSOCS,
and red is the path taken by real robots running NTOC.}
\vspace{-1.5em}
\label{fig:TruePaths}
\end{figure}

We performed ablation tests in simulation to see how different parts of TSOCS affect its success rate. We also performed experiments in simulation and on real robots to evaluate 1)
the execution times for TSOCS, 2) its accuracy at reaching final locations and
velocities, and 3) the tradeoff between location and velocity accuracy for
trajectories with non-zero final velocities. For the experiments
with zero final velocities, we compared the results of TSOCS to NTOC.

\paragraph{TSOCS implementation}
We implemented TSOCS in C++ using Ceres-Solver~\cite{ceres-solver} as the
nonlinear least-squares optimizer. Our implementation took an average of
$0.38\si{\milli\second}$ to solve the TOC-ORBA problem for each iteration of
closed loop control (\secref{iterative}), with some iterations taking as long as $9\si{\milli\second}$. The controller runs at
$60\si{\hertz}$, so our implementation is able to execute within the timing
constraints for real-time control.

\paragraph{Experiments in simulation}
We simulated omnidirectional robots with actuation noise at
every time step such that $v(t) = v_u(t)*\eta(1, n)$ where $v_u$ is the
expected velocity from executing the control and $n$ is the noise level.
%The displacement in each time step is calculated assuming constant
%acceleration during that time step.
TSOCS is run without regularization and with $\beta = 1$ at all
times in simulation. For comparing TSOCS to NTOC with problems with zero final
velocity, we generated 10,000 random problems by sampling initial
locations and velocities of the robot and setting the desired final
location to the origin at rest. For evaluating the performance of TSOCS with
problems with non-zero final velocities, we generated 10,000 problems by
sampling initial locations and velocities, and setting the desired final
location to the origin, with randomly sampled final velocities.

\paragraph{Experiments with real robots}
We use four-wheeled omnidirectional robots designed for use in
the RoboCup Small Size League~\cite{weitzenfeld2014robocup}. For state
estimation, we used SSL-Vision~\cite{zickler2014five} with
ceiling-mounted cameras to track the robots using colored markers, and
performed state estimation with an extended Kalman filter.
We limited the robots to a maximum of 2 m/s$^2$ acceleration.

We generated problems for real robots by starting the robot at the origin with
randomly sampled initial velocities and final locations
within a $2\si{\meter}\times2\si{\meter}$ window of the starting location. We
sampled 20 problems each with zero, and non-zero final velocities, and repeated
each problem 5 times each, for a total of 100 trials for each condition on the
real robot. We compared TSOCS and NTOC for the set of problems with zero final velocity,
and evaluated TSOCS on the separate set of problems with nonzero final
velocity, which NTOC cannot solve. We refer to TSOCS problems with zero final
velocity as TSOCS-r.

\subsection{Ablation Tests}

To compare how different features of TSOCS affect its success rate, we ran one million trials of the TSOCS solver on problems with nonzero final velocity. We performed one iteration of TSOCS per problem, rather than a full control sequence, to test the full two-stage solver. \tabref{AblationTests} shows the results without various features: $T_{\mathrm{max}}$ refers to the upper bound on travel time that stage 1 uses as an initial guess and holds constant. Without $T_{\mathrm{max}}$, we use $T=1$ for stage 1. Without the initial guess described in \secref{LSO}, we guess $\alphas=(1,2,3,4)$.
Stage 1, along with the initial guess for $\alphas$ and $T$, both improve TSOCS' success rate.
TSOCS with the time upper bound, initial guess, and first stage (the configuration used on all subsequent experiments) can solve over 99\% of problems.

\begin{table}
\centering
\begin{tabular}{|l|r|}
  \hline
  \textbf{Solution Method}                            & \multicolumn{1}{|l|}{\textbf{Failure Rate}} \\
  \hline
  No $T_{\mathrm{max}}$, No Initial Guess, No Stage 1 & 30.49\% \\
  \hline
  $T_{\mathrm{max}}$, No Initial Guess, No Stage 1    & 23.53\% \\
  \hline
  No $T_{\mathrm{max}}$, No Initial Guess, Stage 1    & 9.76\% \\
  \hline
  $T_{\mathrm{max}}$, Initial guess, No Stage 1       & 9.25\% \\
  \hline
  $T_{\mathrm{max}}$, Initial Guess, Stage 1          & 0.39\% \\
  \hline
\end{tabular}
\caption{Impact of different steps on TSOCS failure rate.}
\tablabel{AblationTests}
\vspace{-2em}
\end{table}

\subsection{Accuracy in Travel Time}
\vspace{-1em}
\begin{table}[htb]
  \centering
  \small
  \begin{tabular}{|l|r|r|r|r|}
    \hline
    \multirow{2}{*}{\textbf{Control}}
    & \multirow{2}{4em}{\textbf{Real Robots}} & \multicolumn{2}{c|}{\textbf{Simulation}} \\
    \cline{3-4}
    &  & \multicolumn{1}{c|}{\textbf{No Noise}} & \multicolumn{1}{c|}{\textbf{$5\%$  Noise}} \\
%    \cline{2-73-8}
    \hline
    \footnotesize TSOCS   & \footnotesize .13 [.09, .19] & \footnotesize 0.00 [-0.01, 0.00] & \footnotesize 0.09 [-0.09, 0.90] \\\hline
    \footnotesize TSOCS-r & \footnotesize .34 [.13, .57]  & \footnotesize -0.01 [-0.01, 0.00]  & \footnotesize 0.07 [-0.12, 0.99] \\\hline
    \footnotesize NTOC    & \footnotesize .33 [.15, .53]  & \footnotesize 0.00 [0.00, 0.05]  & \footnotesize 0.43 [0.00, 2.01]  \\\hline
  \end{tabular}
  \caption{Median $T_{\mathrm{rel}}$ with 95\% confidence intervals.}
  \tablabel{TimeTakenComparison}
  \vspace{-1.0em}
\end{table}

To compare travel time accuracy between TSOCS and NTOC, we compared
$T_{\mathrm{rel}}=\frac{T_f - T_o}{T_o}$, where $T_f$ is the actual time taken
by the robot to reach its goal and $T_o$ is the optimal time for the problem.
\tabref{TimeTakenComparison} shows the median and 95\% confidence interval
for $T_{\mathrm{rel}}$ for real and simulated robots.

In simulation without noise, iterative TSOCS and NTOC produce travel times that
are close to optimal, as expected. With 5\% simulated noise, both control
algorithms take longer than the optimal time to execute, but TSOCS gets
significantly closer to the optimal time than NTOC. For
problems with non-zero final velocity and with 5\% noise, TSOCS had a
median execution error of $9\%$ of the optimal time.

On real robots, the effect of noisy actuation is much greater than that of 5\% noise in simulation. TSOCS-r and NTOC both took around a third longer than optimal, partially because the robot could take extra time to reapproach the goal location if it overshot it. On TSOCS problems with nonzero final velocity, the robot could not reapproach the goal without incurring a large time penalty, which time regularization (\eqref{iterative-cost}) prevents, so the relative time of problems with nonzero final velocity was lower.

\subsection{Accuracy in Execution}

We ran TSOCS and NTOC on the same problem conditions with zero final velocity,
and separate TSOCS problems with nonzero final velocity that NTOC could
not solve. We show examples of paths taken by TSOCS and NTOC alongside optimal
paths in ~\figref{TruePaths}. For problems with zero final velocity, TSOCS-r and NTOC had similar
final location and velocity errors. All final location errors were less than
$7\si{\milli\meter}$, and all final velocity errors were less than
$100\si{\milli\meter\per\second}$.

\subsection{Tradeoff Between Location and Velocity Error}

In TSOCS problems with nonzero final velocity, there is a tradeoff between
location accuracy and velocity accuracy. Unlike in problems with zero final
velocity, if a robot overshoots its goal, it
cannot immediately re-approach the final location without significant
backtracking. Thus when a robot is perturbed slightly from its
path and can no longer exactly acheive its goal location and velocity without
backtracking, it must compromise by accepting either some location error or some
velocity error. The $\beta_{\mathrm{min}}$ parameter of the iterative cost
function in \eqref{iterative-cost} controls that tradeoff.

\begin{figure}
  \begin{minipage}[b]{.5\linewidth}
    \begin{overpic}[width=\linewidth,unit=1mm]{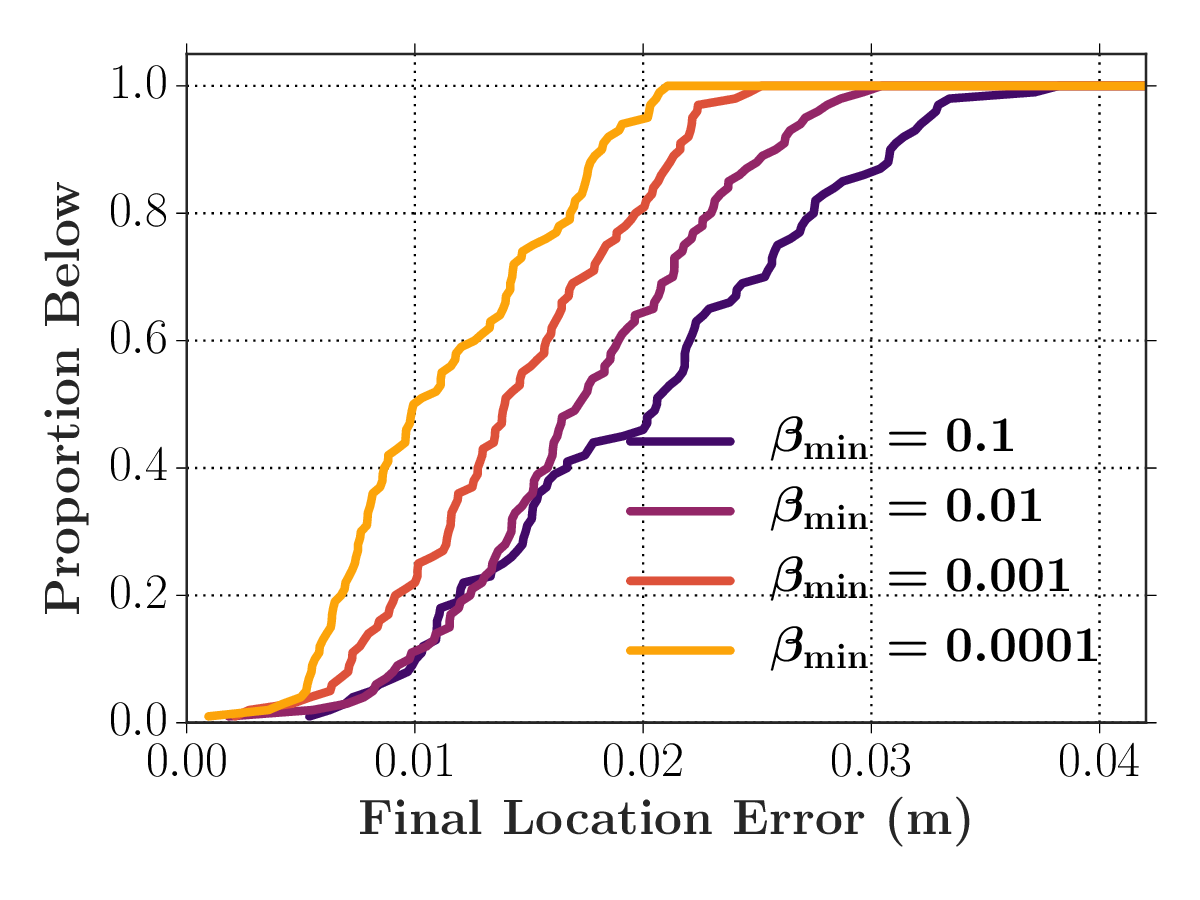}
    \end{overpic}
    \vspace{-0.5cm}
    \subcaption{Location errors.}
  \end{minipage}%
  \begin{minipage}[b]{.5\linewidth}
    \begin{overpic}[width=\linewidth,unit=1mm]{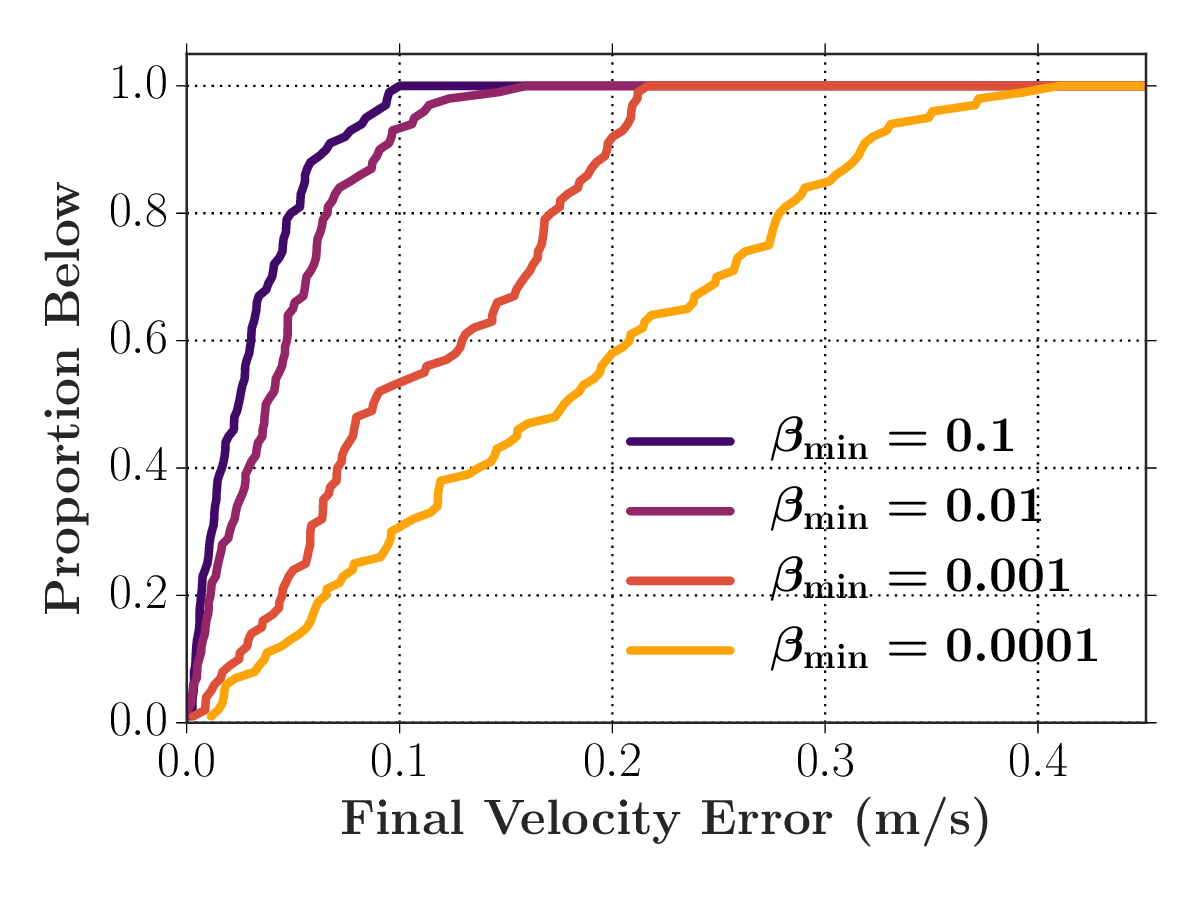}
    \end{overpic}
    \vspace{-0.5cm}
    \subcaption{Velocity errors.}
  \end{minipage}
  \caption{The tradeoff in location accuracy and velocity accuracy in problems with nonzero velocity on real robots.}
  \vspace{-0.75cm}
  \begin{comment} Varying $\beta_{min}$ from \eqref{iterative-cost} allows one to make the robots weight location accuracy and velocity accuracy differently.\end{comment}
  \label{fig:XVTradeoff}
\end{figure}

\figref{XVTradeoff} shows that increasing $\beta_{\mathrm{min}}$, which
increases the weight given to the velocity error in the modified cost function
in~\eqref{iterative-cost}, increases location error and decreases
velocity error, as expected. For $\beta_{\mathrm{min}}=0.01$, the final
velocity errors were less than $0.16\si{\meter\per\second}$, and the final
location errors less than $0.03\si{\meter\per\second}$ for all trials. We used
$\beta_{\mathrm{min}}=0.01$ in all other experiments with TSOCS on real robots.

\begin{comment}
We define the predicted error at any time to be the distances between the location and velocity at end of the trajectory returned by the solver at that time from those of the goal. \figref{PredictedXVError} shows that predicted errors in location and velocity remain zero for most of any problem but increase near its end. This is because the end of any trajectory with nonzero final velocity will feature near one dimensional motion which is particularly sensitive to noise. The small spikes in the predicted error confidence intervals correspond to solver failures; that the predicted error decreases to zero after shows that TSOCS is robust to solver failures and can function adequately by following open loop control until the solver successfully finds a trajectory.
\end{comment}
\begin{figure}
  \begin{minipage}[b]{.5\linewidth}
    \begin{overpic}[width=\linewidth,unit=1mm]{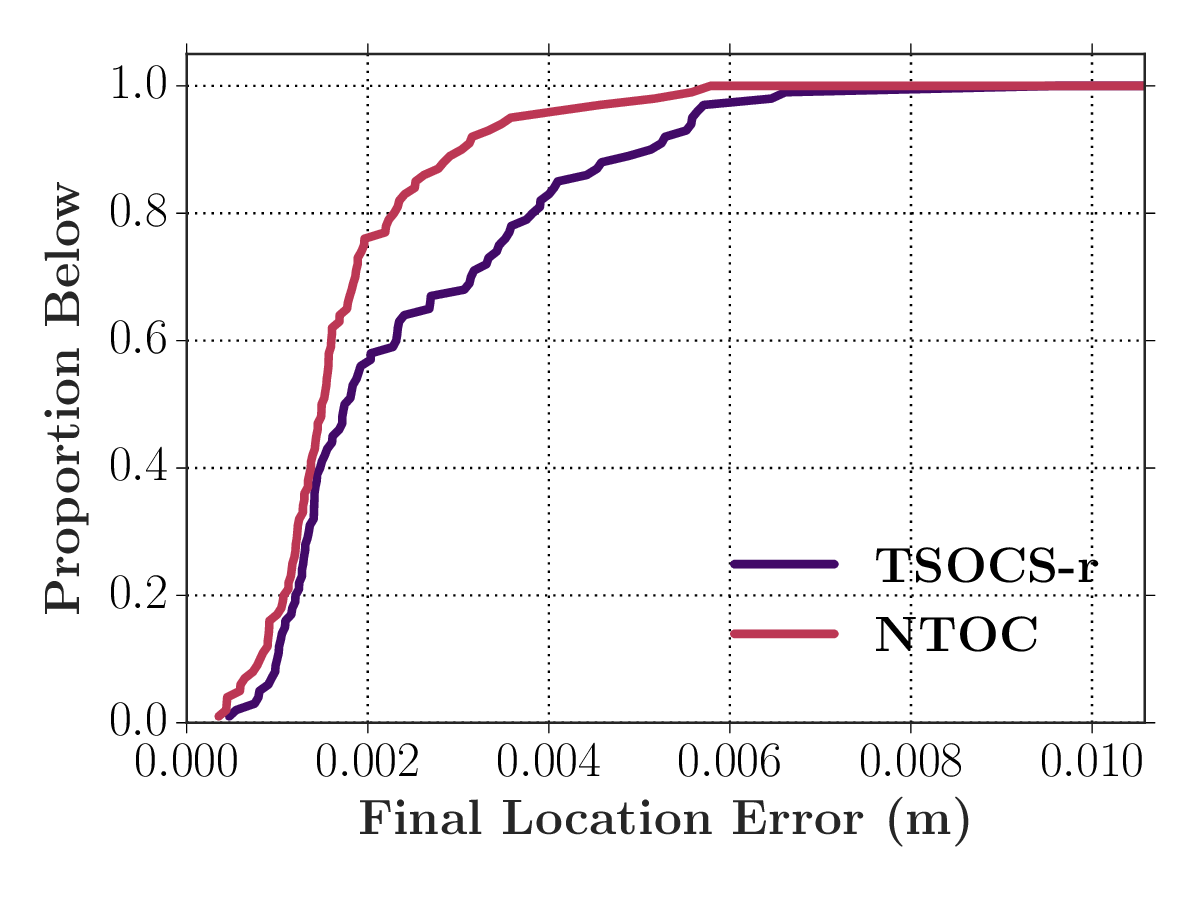}
    \end{overpic}
    \vspace{-0.5cm}
    \subcaption{Location errors.}
  \end{minipage}%
  \begin{minipage}[b]{.5\linewidth}
    \begin{overpic}[width=\linewidth,unit=1mm]{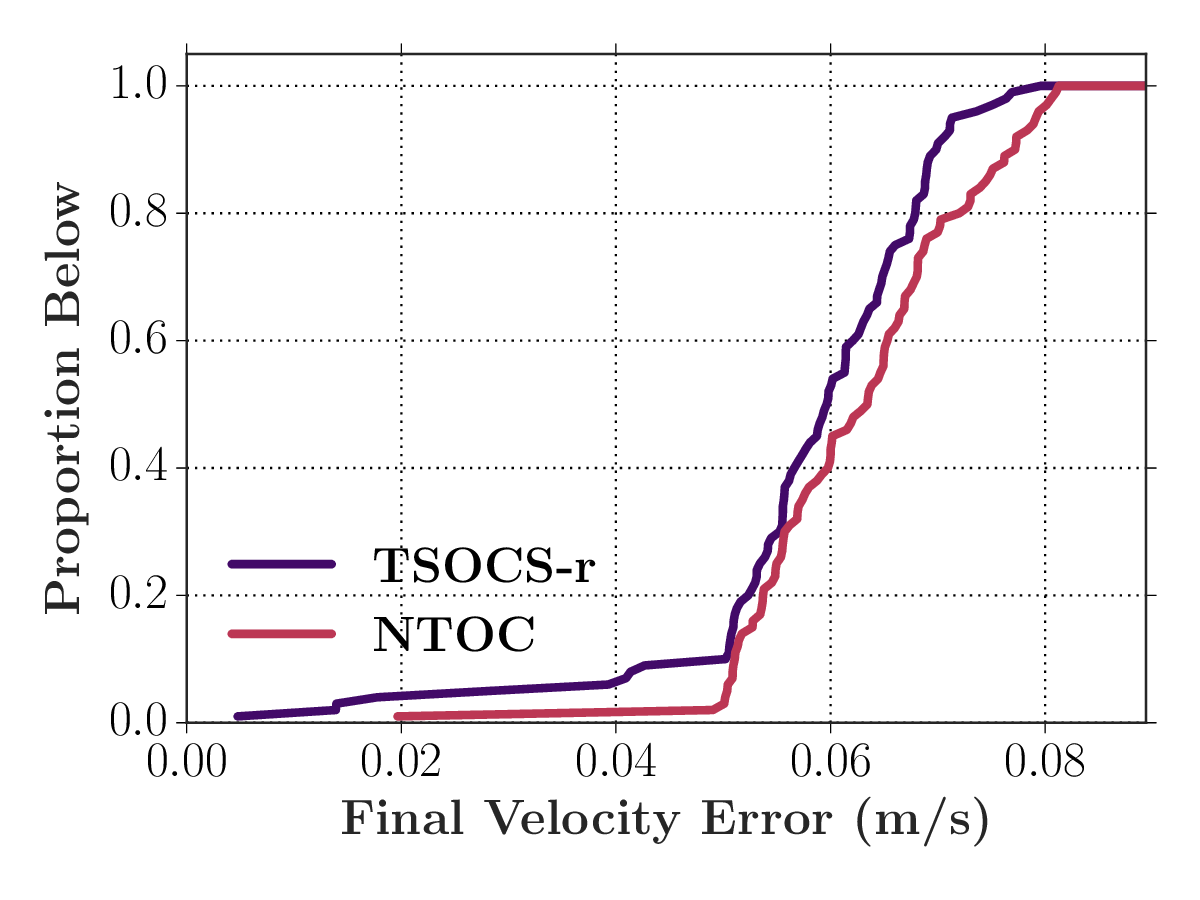}
    \end{overpic}
    \vspace{-0.5cm}
    \subcaption{Velocity errors.}
  \end{minipage}
  \caption{Distributions of TSOCS-r and NTOC errors in final location and velocity on real robots.}
  \label{fig:XVErrorDistributions}
  \vspace{-0.5cm}
\end{figure}

\section{Conclusion}
\seclabel{conclusion}
We introduced a two-stage optimal control solver (TSOCS) to solve the
time-optimal control problem for omnidirectional robots with bounded
acceleration and unbounded velocity. TSOCS can exactly solve problems with both initial and final velocity,
which were previously unsolvable. We demonstrated a closed loop real-time controller using 
TSOCS to control robots with noisy actuation in
simulation, and on real robots. 
% Potential future work includes accounting for
% bounded velocity, theoretical convergence analysis of TSOCS, further evaluation
% on real robots, and applications to other motion tasks such as time optimal ball
% interception for robot soccer.
\begin{comment}
In this paper, we introduced a two-stage optimal control solver (TSOCS) to solve the
time-optimal control problem for omnidirectional robots with bounded acceleration. TSOCS is
capable of successfully solving $>99.9\%$ of problems, as empirically tested over more than
2.9 million problems. TSOCS can solve
problems with both initial and final velocity, which were previously unsolvable in either
near-optimal or optimal solvers under bounded acceleration constraints. We also
demonstrated how TSOCS can be used for closed-loop feedback control under noisy actuation.
Potential future work includes accounting for bounded velocity, theoretical convergence analysis of TSOCS, and further evaluation on real robots.
\end{comment}
\bibliographystyle{IEEEtran}
\bibliography{references}

\begin{thebibliography}{10}
\providecommand{\url}[1]{#1}
\csname url@samestyle\endcsname
\providecommand{\newblock}{\relax}
\providecommand{\bibinfo}[2]{#2}
\providecommand{\BIBentrySTDinterwordspacing}{\spaceskip=0pt\relax}
\providecommand{\BIBentryALTinterwordstretchfactor}{4}
\providecommand{\BIBentryALTinterwordspacing}{\spaceskip=\fontdimen2\font plus
\BIBentryALTinterwordstretchfactor\fontdimen3\font minus
  \fontdimen4\font\relax}
\providecommand{\BIBforeignlanguage}[2]{{%
\expandafter\ifx\csname l@#1\endcsname\relax
\typeout{** WARNING: IEEEtran.bst: No hyphenation pattern has been}%
\typeout{** loaded for the language `#1'. Using the pattern for}%
\typeout{** the default language instead.}%
\else
\language=\csname l@#1\endcsname
\fi
#2}}
\providecommand{\BIBdecl}{\relax}
\BIBdecl

\bibitem{biswas20161000km}
J.~Biswas and M.~Veloso, ``The 1,000-km challenge: Insights and quantitative
  and qualitative results,'' \emph{{ IEEE Intelligent Systems }}, vol.~31,
  no.~3, pp. 86--96, 2016.

\bibitem{mendoza2016selective}
J.~P. Mendoza, J.~Biswas, P.~Cooksey, R.~Wang, S.~Klee, D.~Zhu, and M.~Veloso,
  ``Selectively reactive coordination for a team of robot soccer champions,''
  in \emph{{ AAAI Conference on Artificial Intelligence }}, 2016, pp.
  3354--3360.

\bibitem{rohrig2010localization}
C.~R{\"o}hrig, D.~He{\ss}, C.~Kirsch, and F.~K{\"u}nemund, ``{Localization of
  an omnidirectional transport robot using IEEE 802.15. 4a ranging and laser
  range finder},'' in \emph{IEEE/RSJ International Conference on Intelligent
  Robots and Systems}, 2010, pp. 3798--3803.

\bibitem{pin1994new}
F.~G. Pin and S.~M. Killough, ``A new family of omnidirectional and holonomic
  wheeled platforms for mobile robots,'' \emph{IEEE transactions on robotics
  and automation}, vol.~10, no.~4, pp. 480--489, 1994.

\bibitem{kalmar2004near}
T.~Kalm{\'a}r-Nagy, R.~D’Andrea, and P.~Ganguly, ``Near-optimal dynamic
  trajectory generation and control of an omnidirectional vehicle,''
  \emph{Robotics and Autonomous Systems}, vol.~46, no.~1, pp. 47--64, 2004.

\bibitem{wang2012analytical}
W.~Wang and D.~J. Balkcom, ``Analytical time-optimal trajectories for an
  omni-directional vehicle,'' in \emph{IEEE International Conference on
  Robotics and Automation}, 2012, pp. 4519--4524.

\bibitem{Pifko2008}
S.~Pifko, A.~Zorn, and M.~West, ``Geometric interpretation of adjoint equations
  in optimal low thrust trajectories,'' \emph{Guidance, Navigation, and Control
  and Co-located Conferences}, Aug 2008.

\bibitem{pontryagin1987mathematical}
L.~S. Pontryagin, \emph{Mathematical theory of optimal processes}.\hskip 1em
  plus 0.5em minus 0.4em\relax CRC Press, 1987.

\bibitem{renaud1997minimum}
M.~Renaud and J.-Y. Fourquet, ``Minimum time motion of a mobile robot with two
  independent, acceleration-driven wheels,'' in \emph{IEEE International
  Conference on Robotics and Automation}, 1997, pp. 2608--2613.

\bibitem{chyba1999time}
M.~Chyba and S.~Sekhavat, ``Time optimal paths for a mobile robot with one
  trailer,'' in \emph{IEEE/RSJ International Conference on Intelligent Robots
  and Systems}, vol.~3, 1999, pp. 1669--1674.

\bibitem{verscheure2009time}
D.~Verscheure, B.~Demeulenaere, J.~Swevers, J.~De~Schutter, and M.~Diehl,
  ``Time-optimal path tracking for robots: A convex optimization approach,''
  \emph{IEEE Transactions on Automatic Control}, vol.~54, no.~10, pp.
  2318--2327, 2009.

\bibitem{kuhn1951}
H.~W. Kuhn and A.~W. Tucker, ``Nonlinear programming,'' in \emph{Berkeley
  Symposium on Mathematical Statistics and Probability}.\hskip 1em plus 0.5em
  minus 0.4em\relax Berkeley, Calif.: University of California Press, 1951, pp.
  481--492.

\bibitem{mohammed2016dynamic}
A.~M. Mohammed and S.~Li, ``Dynamic neural networks for kinematic redundancy
  resolution of parallel stewart platforms,'' \emph{IEEE transactions on
  cybernetics}, vol.~46, no.~7, pp. 1538--1550, 2016.

\bibitem{altche2016time}
F.~Altch{\'e}, X.~Qian, and A.~de~La~Fortelle, ``Time-optimal coordination of
  mobile robots along specified paths,'' in \emph{IEEE/RSJ International
  Conference on Intelligent Robots and Systems}, 2016, pp. 5020--5026.

\bibitem{davoodi2013multi}
M.~Davoodi, F.~Panahi, A.~Mohades, and S.~N. Hashemi, ``Multi-objective path
  planning in discrete space,'' \emph{Applied Soft Computing}, vol.~13, no.~1,
  pp. 709--720, 2013.

\bibitem{li2009motion}
X.~Li and A.~Zell, ``Motion control of an omnidirectional mobile robot,'' in
  \emph{Informatics in Control, Automation and Robotics}.\hskip 1em plus 0.5em
  minus 0.4em\relax Springer, 2009, pp. 181--193.

\bibitem{penaloza2015motion}
O.~Pe{\~n}aloza-Mej{\'\i}a, L.~A. M{\'a}rquez-Mart{\'\i}nez, J.~Alvarez, M.~G.
  Villarreal-Cervantes, and R.~Garc{\'\i}a-Hern{\'a}ndez, ``Motion control
  design for an omnidirectional mobile robot subject to velocity constraints,''
  \emph{Mathematical Problems in Engineering}, vol. 2015, 2015.

\bibitem{balkcom2006time}
D.~J. Balkcom, P.~A. Kavathekar, and M.~T. Mason, ``Time-optimal trajectories
  for an omni-directional vehicle,'' \emph{The International Journal of
  Robotics Research}, vol.~25, no.~10, pp. 985--999, 2006.

\bibitem{ceres-solver}
S.~Agarwal, K.~Mierle, and Others, ``Ceres solver,''
  \url{http://ceres-solver.org}.

\bibitem{weitzenfeld2014robocup}
A.~Weitzenfeld, J.~Biswas, M.~Akar, and K.~Sukvichai, ``Robocup small-size
  league: Past, present and future,'' in \emph{Robot Soccer World Cup}.\hskip
  1em plus 0.5em minus 0.4em\relax Springer, 2014, pp. 611--623.

\bibitem{zickler2014five}
Zickler, Laue, G.~Jr, Birbach, Biswas, and Veloso, ``Five years of
  ssl-vision--impact and development,'' in \emph{RoboCup 2013: Robot World Cup
  XVII}.\hskip 1em plus 0.5em minus 0.4em\relax Springer Berlin Heidelberg,
  2014, pp. 656--663.

\end{thebibliography}

\clearpage

\end{document}